\documentclass[letterpaper]{article} 
\usepackage{aaai2027}  
\usepackage[hyphens]{url}  
\usepackage{graphicx} 
\usepackage{natbib}  
\usepackage{caption} 
\usepackage[ruled,linesnumbered]{algorithm2e}
\usepackage{booktabs}
\usepackage{amsmath}
\usepackage{amssymb}
\usepackage{multirow}

\nocopyright 
\usepackage{xspace}

\def\our{\textsc{CSES}\xspace}

\title{Coverage-Driven Adaptive Keyframe Selection for Video Understanding}
\author{Junyang Zhang, Puhan Luo, Chen Tang, Yuxi Shi, Xiang-Yang Li}
\affiliations{University of Science and Technology of China, Hefei, China 
}

\begin{document}

\maketitle

\begin{abstract}
Recent advances in large vision-language models (LVLMs) have enabled long-video understanding and analysis. However, processing the large number of frames in a video incurs substantial computational overhead. Existing methods reduce LVLM inference costs by scoring frame–query relevance before inference and selecting keyframes accordingly. Nevertheless, the distribution of relevant frames varies across queries, and these methods often need to score hundreds or thousands of frames. 
To address this limitation, we propose CSES, a training-free semantic keyframe selector that adaptively determines the numbers of frames to score and keyframes to select.
CSES estimates the prominence of the frame–query relevance profile to guide active acquisition and adapt the temporal coverage of each input. 
It then formulates keyframe selection as a coverage problem that jointly accounts for semantic relevance, temporal redundancy, and visual redundancy. 
Active acquisition and keyframe selection terminate based on coverage saturation.
The selection objective is monotone and submodular, enabling greedy optimization with a standard approximation guarantee.
Experiments with four LVLMs on two benchmarks show that our method preserves accuracy while scoring $4$--$13\times$ fewer frames and selecting $18.4\%$--$20.5\%$ fewer input keyframes than existing baselines.
CSES further achieves a $3.1$--$5.4\times$ speedup in frame selection over baselines.
\end{abstract}


\section{Introduction}
\label{sec:introduction}

Large vision--language models (LVLMs) have enabled the understanding of videos lasting tens of minutes or longer, supporting question answering and reasoning about events, objects, and temporal relationships~\cite{chandr2024hourvide,song2024moviechat}. This capability allows models to combine information from distant parts of a video. However, such videos may contain thousands of frames. Encoding all frames generates a large number of visual tokens, consumes substantial context-window capacity, and increases LVLM inference costs. In addition, irrelevant frames may distract the model from the evidence required to answer a query.

Query-aware keyframe selection addresses this issue by reducing the visual input without modifying the LVLM. A pretrained image--text model first scores each candidate frame for its relevance to the query, after which a selector forwards a selected subset to the frozen LVLM. Training-free methods such as Adaptive Keyframe Sampling (AKS)~\cite{tang2025adaptive} and FOCUS~\cite{zhu2026focus} adaptively determine the temporal locations at which frames are scored or selected under fixed budgets. 
However, scoring hundreds or thousands of candidate frames can itself become a bottleneck. On the same hardware, FOCUS frame selection takes 24.4 seconds per LongVideoBench instance, versus 7.6 seconds for Qwen2-VL-7B inference on its 32 selected frames.
An effective keyframe selection method should therefore adaptively determine the number of frames to score and the number of keyframes to select based on the evidence requirements of each video--query pair. 
The selection criterion should jointly account for query relevance and inter-frame redundancy to preserve relevant evidence while reducing redundant frames.

Jointly adapting these two frame counts is challenging because evidence distributions vary across video--query pairs and must be inferred from sparse observations. 
Sampling densely around observed relevant frames may miss evidence needed for global queries, whereas uniform scoring may overlook short relevant segments and waste computation on irrelevant frames. 
Moreover, relevance scores are unavailable for unscored frames. The selector must therefore infer which frames are worth scoring and adaptively construct a relevant, nonredundant input for the LVLM.

%
In this paper, we formulate keyframe selection as a relevance-weighted coverage problem that jointly considers temporal and visual redundancy.
Redundant frames can be covered by representative frames, while relevance weighting favors frames related to the query.
We observe that queries requiring localized evidence often yield concentrated relevance peaks, whereas those requiring global evidence produce more diffuse relevance distributions. We therefore introduce peak prominence to quantify relevance concentration and adapt the inter-frame temporal coverage range. A smaller range enables denser selection within highly relevant regions, whereas a larger range promotes broader temporal coverage.




We further propose a coverage-driven active acquisition strategy to allocate the scoring budget efficiently. Coarse uniform sampling provides initial relevance observations and estimates relevance concentration. 
The method then prioritizes unscored frames near high-relevance or insufficiently covered regions based on their expected marginal coverage gains.
This prevents scoring operations from being repeatedly concentrated within a small number of temporal segments.
Both active acquisition and final keyframe selection are driven by the coverage principle and terminate based on coverage saturation, allowing the numbers of scored frames and selected keyframes to adapt to each video--query pair.

Our contributions are as follows:
\begin{itemize}
    \item We analyze cross-query variations in frame--query relevance profiles and introduce peak prominence, a measure of relevance concentration that guides adaptive frame selection across video--query pairs.
    \item We propose \our, a training-free selector that uses peak prominence to guide active acquisition and relevance-weighted visual--temporal submodular coverage, thereby adapting both frame counts.
    \item Across four LVLMs and two benchmarks, \our attains comparable observed accuracy while the baselines score $4.1$--$13.0\times$ as many frames and \our selects $18.4\%$--$20.5\%$ fewer input keyframes.
\end{itemize}

\section{Related Work}
\label{sec:related-work}

\textbf{Query-Aware Keyframe Selection.}
Many training-free query-aware selectors use CLIP or BLIP to score frame--query relevance and retain a compact input subset~\cite{radford2021learning,li2022blip}. AKS recursively partitions the timeline to balance relevance and temporal coverage, whereas FOCUS distributes a prescribed scoring budget across temporal arms~\cite{tang2025adaptive,zhu2026focus}. Q-Frame jointly allocates frames and resolution, Logic-in-Frames uses semantic-logical verification, and MDP$^3$ models relevance, diversity, and temporal order under a fixed output budget~\cite{zhang2025qframe,guo2025logic,sun2025mdp3}. Learned selectors use pseudo-labels, downstream feedback, or rewards~\cite{yu2023self,yu2024framevoyager,hu2025mllm,qin2026efficient}. Flexible Frame Selector can also predict a variable output size~\cite{buch2025flexible}. Recent training-free methods adapt sampling to query type or temporal structure~\cite{li2026divide,zhang2026adaq,chen2026wavelet}, while others use query-, content-, or model-derived signals for test-time selection~\cite{peng2026qca,eltahir2026gridprobe,wang2026dafs}. Agent-based systems retrieve evidence over multiple rounds and operate under a different computational setting~\cite{wang2024videoagent,chu2025graphvideoagent,liu2026longvideoagent}. Among pre-inference selectors, joint per-instance control of the realized frame-scoring count and LVLM input size remains underexplored.

\textbf{Coverage-Based Selection.}
Coverage objectives support representative, nonredundant subset selection. Submodular formulations are used in document and video summarization and admit standard greedy guarantees under cardinality constraints~\cite{lin2011class,gygli2015video,nemhauser1978analysis}. Classical formulations assume that candidate utilities are available before selection, whereas scorer-based query-aware methods must first decide which frames to score. \our applies marginal coverage to acquisition under partial observation and final selection over scored frames, while keeping the objectives distinct. Coverage saturation adaptively stops both stages within their respective upper bounds.

\section{Observation}
\label{sec:signal-characterization}


To enable adaptation across different video--query pairs, we first analyze variations in relevance distributions, then define a metric to characterize these variations.

\begin{figure}[t]
    \centering
    \includegraphics[width=\columnwidth]{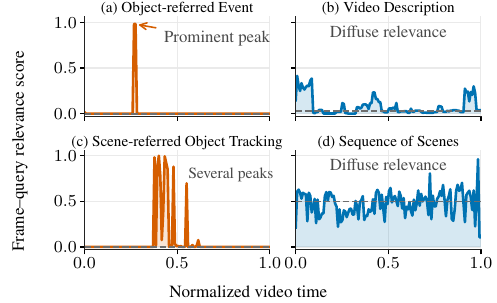}
    \caption{Representative frame--query relevance profiles.}
    \label{fig:relevance-profiles}
\end{figure}


\noindent
\textbf{Relevance distributions vary across video--query pairs.}
Figure~\ref{fig:relevance-profiles} shows BLIP relevance profiles computed from 128 uniformly sampled frames. 
Panels (a), (c), and (d) present examples from LongVideoBench. Figure~\ref{fig:relevance-profiles}(b) uses the same video as (a), but with a query requiring a global description. 
Queries requiring localized evidence tend to produce one or several prominent peaks, whereas queries requiring video-wide or multi-moment evidence produce more diffuse relevance profiles.
These observations indicate that the relevance distribution is jointly determined by the video content and the query, motivating an instance-adaptive selection strategy.
We therefore define log peak prominence to quantify the concentration of each relevance profile.

\begin{figure*}[t]
    \centering
    \begin{minipage}[t]{0.473\textwidth}
        \centering
        \includegraphics[width=\linewidth]{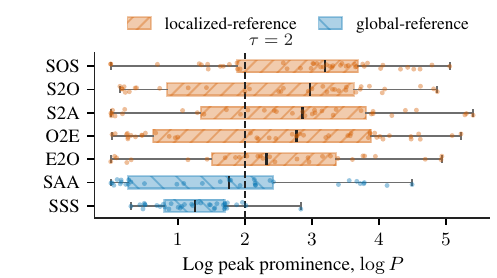}
        \captionof{figure}{Log peak prominence across question types.}
        \label{fig:logp-by-task}
    \end{minipage}
    \hfill
    \begin{minipage}[t]{0.473\textwidth}
        \centering
        \includegraphics[width=\linewidth]{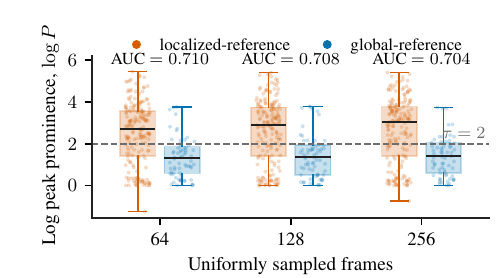}
        \captionof{figure}{Stability across uniform probe sizes.}
        \label{fig:logp-sampling}
    \end{minipage}
\end{figure*}


\noindent
\textbf{Log peak prominence.}
The maximum relevance score alone does not reveal whether the peak is prominent relative to background variation. For a relevance profile $r=(r_1,\ldots,r_n)$, we define peak prominence as
\begin{equation}
P=
\frac{\max(r)-\operatorname{median}(r)}
{\operatorname{MAD}(r)+\delta},
\label{eq:peak-prominence}
\end{equation}
where $\operatorname{MAD}(r)=\operatorname{median}(|r-\operatorname{median}(r)|)$ is the median absolute deviation and $\delta=10^{-6}$ is a small constant to prevent division by zero. 
To compress the wide dynamic range of $P$, we use $\log P=\log_{10}(\max(P,10^{-6}))$, referred to as \emph{log peak prominence}. 
A high $\log P$ indicates that the maximum relevance score stands out clearly from background variation, suggesting that query-relevant evidence is concentrated within localized temporal segments.

\noindent
\textbf{Effectiveness and stability.}
We randomly sample 256 video--query pairs from LongVideoBench, covering seven categories in its official taxonomy: Scene-referred Object Tracking (SOS), Scene-referred Object Existence (S2O), Scene-referred Object Attribute (S2A), Object-referred Event (O2E), Event-referred Object (E2O), Scene-referred Object Attribute Change (SAA), and Sequence of Scenes (SSS).
For analysis, we further group these categories according to their temporal information requirements implied by their task definitions.
The first five typically target temporally localized visual information and form the \emph{localized-reference} group.
By contrast, SAA and SSS explicitly require comparison or ordering across multiple moments and form the \emph{global-reference} group.
For each video--query pair, we compute $\log P$ from the relevance scores of 128 uniformly sampled frames.
As shown in Figure~\ref{fig:logp-by-task}, $\log P$ is generally higher for the localized-reference group than for the global-reference group, indicating that it captures differences in relevance concentration across instances.
Based on this observation, we use $\tau=2$ as an empirical reference value to distinguish relatively concentrated from relatively diffuse relevance profiles.
To evaluate stability, we recompute $\log P$ for the same instances using 64, 128, and 256 uniformly sampled frames.
Figure~\ref{fig:logp-sampling} shows that the between-group separation remains broadly consistent across probe sizes.
This empirical stability is consistent with the non-additive construction of $\log P$.
Thus, $\log P$ provides an instance-level signal of relevance concentration, and serves as the basis for adapting the temporal extent of frame coverage in the subsequent method.

\section{Method Design}
\label{sec:method}

\subsection{Method Overview}


\our (\underline{C}overage-driven \underline{S}ubmodular k\underline{E}yframe \underline{S}election) formulates final keyframe selection using a submodular visual--temporal coverage objective weighted by semantic relevance. It extends the same coverage principle to active acquisition, thereby adapting both the frame-scoring count and the LVLM input size to each video--query pair.

\begin{figure*}[t]
    \centering
    \includegraphics[width=0.99\textwidth]{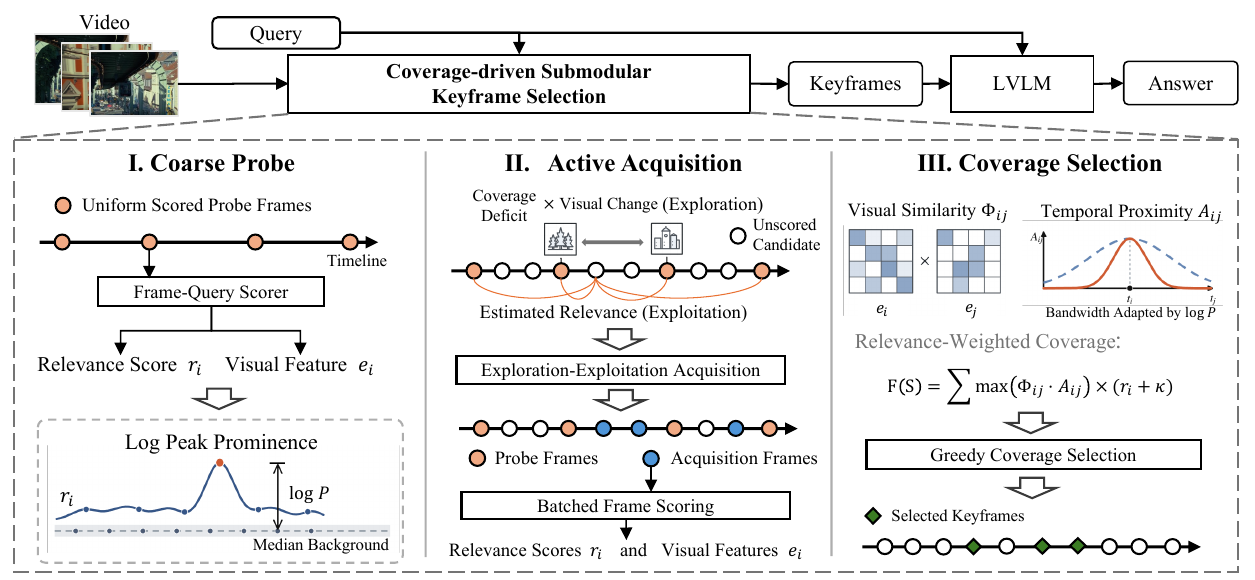}
    \caption{Overview of the three-stage \our framework.}
    \label{fig:framework}
\end{figure*}

As shown in Figure~\ref{fig:framework}, \our proceeds in three stages. First, a coarse probe scores a small set of uniformly sampled frames to estimate relevance concentration. Next, coverage-driven active acquisition prioritizes unscored temporal locations for additional scoring according to their estimated coverage value. Finally, the method uses the relevance scores and visual features of all scored frames to select the keyframe set. The latter two stages assess candidate value through marginal coverage and terminate based on coverage saturation. 
We next formalize the coverage objective underlying \our before presenting the complete three-stage procedure.

\subsection{Relevance-Weighted Visual--Temporal Coverage}
\label{sec:coverage-objective}

We formulate final keyframe selection as a coverage problem: once relevant content has been represented, additional frames that are visually similar to and temporally close to the selected frames offer little marginal gain.
Let $\mathcal O$ be the set of scored frames and $\mathcal S$ the set of selected keyframes. 
For each $i\in\mathcal O$ with timestamp $t_i$, a frame scorer returns a frame--query relevance score $r_i\in[0,1]$ and an $\ell_2$-normalized, query-independent visual feature $e_i$ in one forward pass. The visual feature is obtained from the same image encoding used for relevance scoring and therefore requires no additional image-encoder pass.

We define visual and temporal affinities between a scored frame $i$ and a selected frame $j$, and combine them into a joint coverage kernel $\mathcal K_{ij}$:
\begin{gather}
\Phi_{ij}=\max(\langle e_i,e_j\rangle,0),\;\;
A_{ij}=\exp\!\Big(-\frac{(t_i-t_j)^2}{2\ell^2}\Big),
\label{eq:temporal-kernel}\\
\mathcal K_{ij}=\Phi_{ij}A_{ij},\;\;\;\; \mathcal K_{ii}=1,
\label{eq:joint-kernel}
\end{gather}
where $\Phi_{ij}$ denotes the clipped cosine similarity and captures visual redundancy, whereas $A_{ij}$ is a Gaussian temporal kernel and captures temporal redundancy. 
The parameter $\ell>0$ denotes the temporal bandwidth of $A_{ij}$, measured in seconds.
Both terms are nonnegative, and $\mathcal K_{ij}\in[0,1]$ takes a large value only when the two frames are both visually similar and temporally close. Consequently, visually similar but temporally distant frames and temporally adjacent but visually dissimilar frames are not treated as interchangeable.

The temporal bandwidth $\ell$ determines the temporal range over which one frame can cover another and should adapt to the estimated evidence concentration of each video--query pair. 
A smaller bandwidth preserves fine-grained temporal distinctions around localized evidence, whereas a larger bandwidth suppresses redundant local selection and promotes broader temporal coverage. 
We therefore map the log peak prominence introduced above to a concentration coefficient $u$ and then to $\ell$:
\begin{equation}
u=\operatorname{clip}\!\left(\log P/\tau,0,1\right),\;\;
\ell=1+(1-u)(\ell_{\max}-1),
\label{eq:adaptive-bandwidth}
\end{equation}
where $\ell_{\max}$ is the maximum temporal bandwidth, and $\tau=2$ is the empirical reference value identified in Section~\ref{sec:signal-characterization}. Clipping $\log P/\tau$ to $u\in[0,1]$ limits the influence of extreme values and keeps $\ell\in[1,\ell_{\max}]$ seconds. 
A larger $u$ indicates more concentrated relevant evidence and narrows the temporal kernel, enabling denser selection within localized high-relevance regions. Conversely, a smaller $u$ widens the kernel, discouraging repeated local selection and promoting broader temporal coverage.

The joint kernel captures inter-frame representativeness, but not whether the represented content is useful for the query. We therefore define the relevance-weighted coverage objective:
\begin{equation}
F(\mathcal S)=\sum_{i\in\mathcal O}(r_i+\kappa)
\max_{j\in\mathcal S}\mathcal K_{ij},
\qquad \max_{j\in\varnothing}\mathcal K_{ij}:=0,
\label{eq:coverage-objective}
\end{equation}
where $\kappa>0$ is a small constant that ensures the objective is equivalent to unweighted coverage when relevance scores are zero.
For each frame $i$, $\max_{j\in\mathcal S}\mathcal K_{ij}$ is its coverage by $\mathcal S$, weighted by relevance. The marginal gain therefore favors candidates that represent high-relevance content not yet covered, while visual and temporal redundancy discount repeated selection. Thus, $F$ combines relevance and both forms of redundancy through marginal coverage.



As a nonnegative weighted maximum-coverage function, $F$ can be shown to be normalized, monotone, and submodular. Therefore, the standard greedy algorithm achieves a $(1-1/e)$ approximation guarantee~\cite{nemhauser1978analysis}.
\our follows this greedy approach, and uses coverage saturation to control the stopping condition.
Let $K_{\max}$ denote the maximum number of selected keyframes, and let $\epsilon\in(0,1)$ be the saturation tolerance. 
Starting with $\mathcal S=\varnothing$, the selector repeatedly adds the frame with the largest marginal gain in $F$. Since $\mathcal K_{ii}=1$, the maximum attainable coverage is $F(\mathcal O)=\sum_{i\in\mathcal O}(r_i+\kappa)$.
The selector terminates when $F(\mathcal S)\ge(1-\epsilon)F(\mathcal O)$ or $|\mathcal S|=K_{\max}$. 
This rule adapts the number of selected keyframes to each video--query pair. Localized evidence may reach coverage saturation with a small representative set, whereas evidence distributed throughout the video may require more keyframes, subject to the upper bound $K_{\max}$.
The appendix provides the submodularity proof and detailed selection pseudocode.

\subsection{Three-Stage Keyframe Selection}
\label{sec:three-stage}

The coverage objective requires the relevance score and visual feature of each candidate frame, but these quantities are unknown before scoring. \our must therefore decide which frames to score. Let  $R_{\max}$ and $K_{\max}$ cap the sizes of $\mathcal O$ and $\mathcal S$, respectively. 
Sparse uniform scoring may miss relevant evidence, whereas dense scoring may negate the intended efficiency gains.
As shown in Algorithm~\ref{alg:cses}, 
the resulting pipeline comprises three stages: initial probing to obtain sparse observations, active acquisition to allocate the scoring budget, and final keyframe selection using relevance-weighted visual--temporal coverage.


\noindent\textbf{Stage 1: Coarse Probe.}
\our uniformly samples $n_1$ frames to form the initial scored set $\mathcal O$ and obtain the relevance scores and visual features $\{(r_i,e_i)\}_{i\in\mathcal O}$.
The method then computes $\log P$ using Equation~\eqref{eq:peak-prominence} and obtains the concentration coefficient $u$ and temporal bandwidth $\ell$ from Equation~\eqref{eq:adaptive-bandwidth}. 
With a small number of scoring operations, the coarse probe provides the adaptive parameters and initial coverage information required for subsequent active acquisition and final keyframe selection.



\begin{algorithm}[t]
\caption{\our: three-stage adaptive selection.}
\label{alg:cses}
\small
\KwIn{Budgets $K_{\max}$ and $R_{\max}$; grid size $G$; probe size $n_1$; bandwidth limit $\ell_{\max}$; saturation threshold $\epsilon$}
\KwOut{Keyframe set $\mathcal S$}
\textbf{Stage 1: Coarse Probe}\;
Uniformly sample and score $n_1$ frames, obtaining $\mathcal O$, $r^{(n_1)}$, and $\{r_i,e_i\}_{i\in\mathcal O}$\;
Compute $\log P$ using Eq.~\eqref{eq:peak-prominence}, and obtain $u$, $\ell$ from Eq.~\eqref{eq:adaptive-bandwidth}\;
\textbf{Stage 2: Active Acquisition}\;
Construct $\mathcal G$ from $G$ uniformly spaced unscored frames; compute $A$ using Eq.~\eqref{eq:temporal-kernel}\;
Compute $\hat r(c)$, $\mathrm{vis}(c)$, and initial $\mathrm{cov}_v(c)$ using Eqs.~\eqref{eq:relevance-prediction} and~\eqref{eq:exploration}; set $\mathcal P\leftarrow\varnothing$\;
\While{$|\mathcal O|+|\mathcal P|<R_{\max}$ and $\mathcal G\setminus\mathcal P\ne\varnothing$}{
    $\sigma(c)\leftarrow(1-\mathrm{cov}_v(c))\mathrm{vis}(c)$; set $W(c)$ using Eq.~\eqref{eq:acquisition-weight}\;
    \If{$\sum_{c\in\mathcal G} W(c)\mathrm{cov}_v(c)\ge(1-\epsilon)\sum_{c\in\mathcal G} W(c)$}{
        \textbf{break}\;
    }
    $\Delta(x)\leftarrow\sum_{c\in\mathcal G}W(c)[A_{xc}-\mathrm{cov}_v(c)]_+$ \tcp*{define $[\,\cdot\,]_+=\max(\,\cdot\,,0)$}
    $x^\star\leftarrow\arg\max_{x\in\mathcal G\setminus\mathcal P}\Delta(x)$\;
    \If{$\Delta(x^\star)\le0$}{
        \textbf{break}\;
    }
    $\mathcal P\leftarrow\mathcal P\cup\{x^\star\}$;
    $\mathrm{cov}_v(c)\leftarrow\max(\mathrm{cov}_v(c),A_{x^\star c})$ for $c\in\mathcal G$\;
}
Score the planned frames in $\mathcal P$ with batched inference;
$\mathcal O\leftarrow\mathcal O\cup\mathcal P$\;
\textbf{Stage 3: Coverage Selection}\;
Compute the recalibrated $\log P^{\dagger}$ using Eq.~\eqref{eq:recalibration}\;
Obtain $\ell^{\dagger}$ from Eq.~\eqref{eq:adaptive-bandwidth}\;
Apply \textsc{CoverageSelection} with $\ell^{\dagger}$ to obtain $\mathcal S$\;
\Return $\operatorname{sort}(\mathcal S)$\;
\end{algorithm}

\noindent\textbf{Stage 2: Active Acquisition.}
Stage~2 allocates the remaining scoring budget under partial observation. 
Because relevance scores and visual features are unavailable at unscored frames, the joint coverage objective cannot yet be evaluated.
Instead, \our estimates the potential value of unscored candidates from the coarse-probe observations and selects frames based on estimated temporal coverage.
Moreover, \our adaptively balances exploitation and exploration according to the concentration of the coarse-probe relevance profile.



We select a set $\mathcal G$ of $G$ uniformly spaced, unscored candidate frames along the video timeline. For each $c\in\mathcal G$, let $t_c$ denote its timestamp. 
Using the bandwidth $\ell$ obtained in Stage~1, all temporal affinities $A$ can be computed from timestamps without frame scoring.

We first construct the exploitation signal to prioritize regions surrounding observed high-relevance frames.
Temporal kernel regression then estimates the relevance of each candidate from the scored frames as the exploitation signal:
\begin{equation}
\hat r(c)=
\frac{\sum_{j\in\mathcal O}A_{cj}r_j}
{\sum_{j\in\mathcal O}A_{cj}}.
\label{eq:relevance-prediction}
\end{equation}


We next construct the exploration signal. A candidate is valuable for exploration when it is poorly covered by scored frames and lies in an interval that may contain unobserved information. We capture these two factors by defining the exploration signal $\sigma(c)$:
\begin{equation}
\begin{aligned}
\mathrm{cov}_v(c)
&=\max_{j\in\mathcal O}A_{cj},\\
\mathrm{vis}(c)
&=1-\max\left(\langle e_{j_{\mathrm L}(c)},
e_{j_{\mathrm R}(c)}\rangle,0\right),\\
\sigma(c)
&=(1-\mathrm{cov}_v(c))\mathrm{vis}(c),
\end{aligned}
\label{eq:exploration}
\end{equation}
where $j_{\mathrm L}(c)$ and $j_{\mathrm R}(c)$ denote the probe frames with the nearest timestamps before and after $t_c$. 
$(1-\mathrm{cov}_v(c))$ measures the temporal coverage deficit at $c$ and is large in temporal gaps poorly represented by the scored frames.
$\mathrm{vis}(c)$ measures the visual difference between its neighboring probes and indicates whether the intervening interval may contain unobserved information. Consequently, $\sigma(c)$ is large when $c$ is poorly covered and its neighboring probes differ visually.

We combine these two signals using a concentration coefficient to determine the potential value of candidate frames:
\begin{equation}
W(c)=u\bigl(\hat r(c)+\kappa\bigr)
+(1-u)\sigma(c).
\label{eq:acquisition-weight}
\end{equation}
A higher $u$ value implies that evidence is relatively concentrated, causing active acquisition to focus on the neighborhood of high-relevance frames; a lower $u$ shifts emphasis toward poorly covered intervals with visual change.

To avoid repeatedly scoring redundant frames within local regions, \our selects frames for scoring according to their marginal coverage gains.
Because visual features are unavailable for unscored candidates, the method uses $W(c)$ to weight temporal coverage and greedily selects candidates according to their current weighted marginal temporal-coverage gains (Algorithm~\ref{alg:cses}, lines~7--18). 
Frame acquisition terminates once coverage reaches the prescribed saturation threshold.
The resulting acquisition set $\mathcal P$ is then scored in batches and added to the scored set $\mathcal O$.



\begin{table*}[t]
\centering
\small
\caption{Frame-scoring count, LVLM input size and accuracy. \our reduces both counts while achieving similar mean accuracy.}
\label{tab:main-results}
\begin{tabular*}{\textwidth}{@{\extracolsep{\fill}}l*{10}{c}@{}}
\multicolumn{11}{@{}l}{\textit{(a) Frame-scoring count and LVLM input size}} \\
\toprule
& \multicolumn{6}{c}{Number of scored frames} & \multicolumn{4}{c}{Number of keyframes} \\
\cmidrule(lr){2-7}\cmidrule(l){8-11}
Selector & \multicolumn{2}{c}{$\bar R$} & \multicolumn{2}{c}{Reduction vs. AKS} & \multicolumn{2}{c}{Reduction vs. FOCUS} & \multicolumn{2}{c}{$\overline{|\mathcal S|}$} & \multicolumn{2}{c}{Reduction vs. FOCUS} \\
\cmidrule(lr){2-3}\cmidrule(lr){4-5}\cmidrule(lr){6-7}\cmidrule(lr){8-9}\cmidrule(l){10-11}
& LVB & VMME & LVB & VMME & LVB & VMME & LVB & VMME & LVB & VMME \\
\midrule
AKS & 746.6 & 1039.6 & -- & -- & -- & -- & 32 & 32 & -- & -- \\
FOCUS & 323.4 & 449.4 & $\downarrow 2.3\times$ & $\downarrow 2.3\times$ & -- & -- & 32 & 32 & -- & -- \\
\makebox[0pt][l]{\color{black!4}\rule[-0.3em]{\textwidth}{1.25em}}\our (our) &
\textbf{79.3} &
\textbf{79.9} &
$\boldsymbol{\downarrow 9.4\times}$ &
$\boldsymbol{\downarrow 13.0\times}$ &
$\boldsymbol{\downarrow 4.1\times}$ &
$\boldsymbol{\downarrow 5.6\times}$ &
\textbf{25.4} &
\textbf{26.1} &
$\boldsymbol{\downarrow 20.5\%}$ &
$\boldsymbol{\downarrow 18.4\%}$ \\
\bottomrule
\end{tabular*}

\smallskip

\begin{tabular*}{\textwidth}{@{\extracolsep{\fill}}l*{9}{c}@{}}
\multicolumn{10}{@{}l}{\textit{(b) Accuracy (\%)}} \\
\toprule
& \multicolumn{3}{c}{LVB} & \multicolumn{3}{c}{VMME} & \multicolumn{3}{c}{Mean accuracy} \\
\cmidrule(lr){2-4}\cmidrule(lr){5-7}\cmidrule(l){8-10}
LVLM & AKS & FOCUS & \our (our) & AKS & FOCUS & \our (our) & AKS & FOCUS & \our (our) \\
\midrule
LLaVA-OneVision-7B & 58.19 & \textbf{59.84} & 59.61 & \textbf{60.63} & 58.63 & 60.26 & 59.41 & 59.23 & \textbf{59.93} \\
Qwen2-VL-7B-Instruct & 57.52 & \textbf{59.69} & 58.34 & \textbf{60.00} & 59.48 & \textbf{60.00} & 58.76 & \textbf{59.59} & 59.17 \\
LLaVA-Video-7B & 59.84 & 60.28 & \textbf{60.96} & \textbf{64.26} & 62.56 & 62.96 & \textbf{62.05} & 61.42 & 61.96 \\
Qwen3-VL-8B-Instruct & 58.79 & \textbf{60.51} & 59.54 & \textbf{66.33} & 65.30 & 64.93 & 62.56 & \textbf{62.90} & 62.23 \\
\midrule
Four-model mean & 58.59 & \textbf{60.08} & 59.61 & \textbf{62.81} & 61.49 & 62.04 & 60.70 & 60.79 & \textbf{\underline{60.82}} \\
\bottomrule
\end{tabular*}
\end{table*}

\begin{figure*}[htbp]
    \centering
    \includegraphics[width=\textwidth]{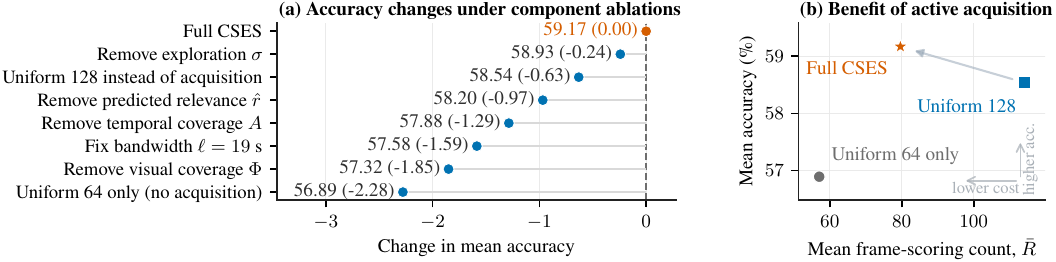}
    \caption{Component ablations with Qwen2-VL-7B-Instruct and two benchmarks.}
    \label{fig:ablation-tradeoff}
\end{figure*}

\noindent\textbf{Stage 3: Coverage Selection.}
Stage~3 applies the final coverage objective to the expanded scored set. 
We refer to the greedy coverage procedure described in Section~\ref{sec:coverage-objective} as \textsc{CoverageSelection}.

Active acquisition may find high-relevance frames missed by the coarse probe, causing the Stage~1 estimate of $\log P$ to underestimate relevance concentration. 
However, active acquisition preferentially samples regions of high predicted relevance, so the expanded scored set is biased and may distort the background median and median absolute deviation when used to recompute $\log P$.
We therefore use Stage~2 scores only to update the peak value while retaining background statistics from the uniform Stage~1 probe. 
Let $r^{(\mathcal O)}$ denote relevance scores over the expanded scored set after Stage~2 and $r^{(n_1)}$ those of the $n_1$ frames scored in Stage~1. We define the recalibrated $P$ as
\begin{equation}
P^{\dagger}
=
\frac{\max(r^{(\mathcal{O})})-\operatorname{median}(r^{(n_1)})}
{\operatorname{MAD}(r^{(n_1)})+\delta}
.
\label{eq:recalibration}
\end{equation}
We then obtain $\log P^{\dagger}$ from $P^{\dagger}$ using the same transformation as in Section~\ref{sec:signal-characterization}.
This hybrid estimate incorporates a newly discovered peak without using the biased acquisition set to estimate the background. Substituting $\log P^{\dagger}$ into Equation~\eqref{eq:adaptive-bandwidth} gives $u^{\dagger}$ and $\ell^{\dagger}$. We then construct the joint kernel over $\mathcal O$ using $\ell^{\dagger}$ and apply \textsc{CoverageSelection}.
Coverage saturation determines the realized size of $\mathcal S$, which is returned in temporal order (Algorithm~\ref{alg:cses}, lines~21--24).



\noindent\textbf{Inference cost.}
Active acquisition and coverage selection follow the coverage principle to adapt two distinct costs to each video--query pair. 
Active acquisition determines the number of additional frames to score subject to $R_{\max}$. 
The resulting frame-scoring count is $R=|\mathcal O|=n_1+|\mathcal P|$.
Coverage selection determines the LVLM input size, producing $|\mathcal S|$ keyframes subject to $K_{\max}$.




\begin{figure*}[t]
    \centering
    \includegraphics[width=\textwidth]{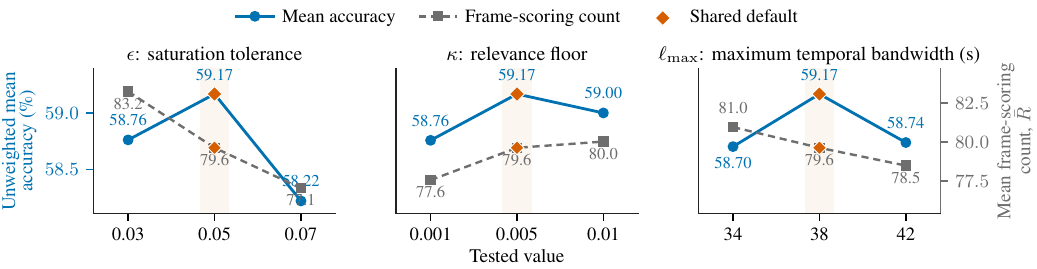}
    \caption{Local parameter sensitivity with Qwen2-VL-7B-Instruct and two benchmarks.}
    \label{fig:parameter-sensitivity}
\end{figure*}

\section{Experiments}
\label{sec:experiments}


\subsection{Experimental Setup}
\label{sec:setup}

\paragraph{Benchmarks and models.}
We use the 1,337-question validation set of LongVideoBench (LVB)~\cite{wu2024longvideobench} and 2,700 Video-MME (VMME) questions~\cite{fu2024videomme}. The downstream LVLMs are LLaVA-OneVision-7B~\cite{li2024llavaonevision}, Qwen2-VL-7B-Instruct~\cite{wang2024qwen2vl}, LLaVA-Video-7B~\cite{zhang2024llavavideo}, and Qwen3-VL-8B-Instruct~\cite{bai2025qwen3vl}.

\paragraph{Baselines.}
We compare with AKS~\cite{tang2025adaptive}, which balances relevance and temporal coverage, and FOCUS~\cite{zhu2026focus}, which allocates samples across temporal arms. All methods are evaluated with LMMs-Eval~\cite{zhang2024lmms} in the same environment. 
They share the BLIP-large frame scorer, $K_{\max}=32$, the prompts, and all downstream LVLM inference settings.

\paragraph{Metrics and configuration.}
We report accuracy, the mean number of scored frames $\bar R$, and the mean number of keyframes forwarded to the LVLM $\overline{|\mathcal S|}$. In our protocol, selector outputs do not depend on the downstream LVLM, so both frame counts are shared across the four models. \our uses $K_{\max}=32$, $R_{\max}=128$, $G=512$, $n_1=64$, $\ell\in[1,38]$ seconds, $\kappa=0.005$ and $\epsilon=0.05$. 
Unless otherwise specified in the ablation studies, we use this configuration for all models and benchmarks.

\subsection{Overall Results}
\label{sec:main-results}

As shown in Table~\ref{tab:main-results}, \our uses substantially fewer scored and forwarded frames, while the three selectors have similar aggregate observed accuracy. The aggregate over the eight model--benchmark combinations is 60.82\% for \our, 60.79\% for FOCUS, and 60.70\% for AKS, and no selector has the highest accuracy in every setting. 

The efficiency gains are consistent across both benchmarks. Relative to \our, AKS scores $9.4$--$13.0\times$ as many frames and FOCUS scores $4.1$--$5.6\times$ as many. \our also forwards $18.4\%$--$20.5\%$ fewer keyframes than their fixed 32-frame outputs. Thus, \our reaches a lower-cost operating point with close aggregate observed accuracy.

\subsection{Method Analysis}
\label{sec:ablation}



\paragraph{Component ablations.}
Figure~\ref{fig:ablation-tradeoff}(a) reports component-wise ablations of \our. Every variant reduces mean accuracy, indicating that each component contributes to the complete method. The largest observed decrease occurs when active acquisition is removed and only the coarse probe is retained, reducing mean accuracy by 2.28 points. To further evaluate active acquisition, we replace it with uniform scoring of up to 128 frames. As shown in Figure~\ref{fig:ablation-tradeoff}(b), active acquisition yields 0.63 points higher mean accuracy while reducing $\bar R$ from 114.1 to 79.6. 
This result indicates that active acquisition allocates the scoring budget more effectively by prioritizing candidate frames with greater estimated value for final keyframe selection.


\paragraph{Coverage selection.}
As shown in Figure~\ref{fig:ablation-tradeoff}(a), fixing $\ell$ at 19 seconds, near the midpoint of its adaptive range, or removing either visual or temporal coverage lowers mean accuracy. These results support the contributions of bandwidth adaptation and joint visual--temporal coverage. To evaluate final selection independently of active acquisition, we apply the Stage~3 coverage selector and AKS to the same 128 uniformly scored candidate frames. Table~\ref{tab:coverage-only} shows that our selector forwards fewer keyframes and achieves a higher mean accuracy than AKS. 
Compared with the 32-frame uniform input, the Stage~3 coverage selector improves mean accuracy while forwarding fewer frames on both benchmarks, showing that a smaller LVLM input does not compromise accuracy.
These results support the effectiveness of the proposed coverage selection method. 
FOCUS is excluded because its selection is coupled to its arm-search procedure.

\begin{table}[t]
\centering
\small
\caption{Comparison of final-selection methods.}
\label{tab:coverage-only}
\begin{tabular}{@{}lcccc@{}}
\toprule
Selector & Frames & LVB & VMME & Mean \\
\midrule
Uniform input & 32 & 55.05 & 57.70 & 56.38 \\
AKS from 128 & 32 & 55.12 & \textbf{59.41} & 57.26 \\
\our from 128 & 26.3/27.6 & \textbf{58.04} & 59.04 & \textbf{58.54} \\
\bottomrule
\end{tabular}
\end{table}


\paragraph{Parameter sensitivity.}
Figure~\ref{fig:parameter-sensitivity} shows limited sensitivity around the default configuration. Across the tested one-at-a-time settings, mean accuracy varies by at most 0.95 points and $\bar R$ ranges from 77.1 to 83.2. The saturation tolerance $\epsilon$ produces the clearest accuracy--cost trade-off, consistent with its role in stopping both coverage stages, whereas $\kappa$ and $\ell_{\max}$ have smaller local effects. 

\begin{table}[t]
\centering
\small
\setlength{\tabcolsep}{3pt}
\caption{Frame-selection latency on LVB (s/instance).}
\label{tab:frame-selection-latency}
\begin{tabular}{@{}lrrrrr@{}}
\toprule
Method & Decode & Preproc. & Inference & Selection & Total \\
\midrule
\our(our) & \textbf{5.052} & \textbf{0.428} & \textbf{2.298} & \textbf{0.088} & \textbf{7.866} \\
AKS & 17.086 & 4.005 & 21.648 & 0.097 & 42.838 \\
FOCUS & 13.156 & 1.746 & 9.372 & 0.138 & 24.411 \\
\bottomrule
\end{tabular}
\end{table}

\subsection{Frame-Selection Latency}
\label{sec:latency}

We measure selector-side latency from video decoding through final selection on all 1,337 LVB validation instances. 
Experiments use an NVIDIA RTX 3090, an Intel Core i5-13600KF, and BLIP-large with a batch size of 8. 
The reported inference latency of \our includes the additional cost of $\ell_2$-normalizing the visual features produced by BLIP.
Table~\ref{tab:frame-selection-latency} reports mean latency per instance for decoding, BLIP preprocessing and GPU transfer, BLIP inference, and the remaining selection overhead.

From Table~\ref{tab:frame-selection-latency}, \our achieves a $5.45\times$ frame-selection speedup over AKS and a $3.10\times$ speedup over FOCUS. Most of the measured reduction occurs in video decoding, BLIP preprocessing, and BLIP inference, consistent with processing fewer frames. For context, on the same hardware, Qwen2-VL-7B-Instruct inference using the 32 keyframes selected by FOCUS averages 7.618 seconds per instance, whereas FOCUS requires 24.411 seconds for frame selection. Thus, scoring many candidate frames with BLIP can make frame selection more costly than downstream LVLM inference. \our reduces this cost to 7.866 seconds, substantially alleviating the bottleneck.




\FloatBarrier



\section{Conclusion}
\label{sec:conclusion}


We presented \our, a training-free keyframe selector that adapts both the frame-scoring count and the LVLM input size. 
\our uses log peak prominence to characterize relevance concentration. 
CSES uses this signal to adapt temporal coverage and guide active acquisition.
Final keyframes are then selected using relevance-weighted visual--temporal submodular coverage.
Across four LVLMs on LongVideoBench and Video-MME, \our achieves comparable observed accuracy while the baselines score $4.1$--$13.0\times$ as many frames and \our forwards $18.4\%$--$20.5\%$ fewer keyframes. 

\bibliography{aaai2027}

\clearpage
\appendix

\section{Additional Details for the Observation}
\label{app:relevance-profile-queries}

The four profiles in Figure~\ref{fig:relevance-profiles} are computed from 128 uniformly sampled frames. Panels (a), (c), and (d) correspond to LongVideoBench samples. Panel (b) uses the same video as panel (a) with an author-constructed video-description query, providing a controlled comparison of different information needs for the same video. The queries and corresponding log peak prominence values are listed below.

\noindent\textbf{(a) Object-referred Event, $\log P=5.22$.}
\textit{Query:} On a stage with lights, there are many people wearing colorful outfits. What are these people in the colorful outfits doing?

\noindent\textbf{(b) Video Description, $\log P=1.20$.}
\textit{Query:} What does this video cover?

\noindent\textbf{(c) Scene-referred Object Tracking, $\log P=5.05$.}
\textit{Query:} In an indoor basketball court with red walls and a yellow floor, there is a girl wearing a purple short-sleeve shirt with her hair tied up, holding a basketball. In which of the following places has the girl appeared?

\noindent\textbf{(d) Sequence of Scenes, $\log P=0.69$.}
\textit{Query:} Which of the following sequence of scenes is correct?

Panels (a) and (c), which require temporally localized evidence, have substantially higher log peak prominence than panels (b) and (d), which require video-wide or multi-moment information. The controlled comparison between panels (a) and (b) further shows that the relevance profile depends jointly on the video and the query.

\begin{table*}[htbp]
\centering
\small
\caption{Frame-selection latency on LongVideoBench, grouped by video duration.}
\label{tab:appendix-frame-selection-latency}
\renewcommand{\arraystretch}{1.03}
\begin{tabular}{@{}llrrrrr@{}}
\toprule
Video duration & Method
& \shortstack{Video\\decoding (s)}
& \shortstack{BLIP\\preprocessing (s)}
& \shortstack{BLIP\\inference (s)}
& \shortstack{Selection\\overhead (s)}
& Total (s) \\
\midrule
\multirow{3}{*}{0--5 min}
& \our  & 1.669 & 0.206 & 1.096 & 0.033 & \textbf{3.003} \\
& AKS   & 1.945 & 0.456 & 2.440 & 0.030 & 4.870 \\
& FOCUS & 2.293 & 0.302 & 1.592 & 0.029 & 4.217 \\
\midrule
\multirow{3}{*}{5--10 min}
& \our  & 6.903 & 0.541 & 2.902 & 0.105 & \textbf{10.452} \\
& AKS   & 11.517 & 2.586 & 13.802 & 0.106 & 28.012 \\
& FOCUS & 8.719 & 1.132 & 5.989 & 0.094 & 15.934 \\
\midrule
\multirow{3}{*}{10--20 min}
& \our  & 7.050 & 0.550 & 2.946 & 0.101 & \textbf{10.646} \\
& AKS   & 25.277 & 5.635 & 30.306 & 0.119 & 61.337 \\
& FOCUS & 18.558 & 2.394 & 12.809 & 0.178 & 33.939 \\
\midrule
\multirow{3}{*}{20--30 min}
& \our  & 7.216 & 0.574 & 3.116 & 0.127 & \textbf{11.033} \\
& AKS   & 33.367 & 7.645 & 41.757 & 0.153 & 82.922 \\
& FOCUS & 24.633 & 3.228 & 17.553 & 0.288 & 45.701 \\
\midrule
\multirow{3}{*}{30--60 min}
& \our  & 7.242 & 0.617 & 3.296 & 0.200 & \textbf{11.354} \\
& AKS   & 53.663 & 13.926 & 75.530 & 0.243 & 143.361 \\
& FOCUS & 42.100 & 5.915 & 31.836 & 0.405 & 80.256 \\
\bottomrule
\end{tabular}
\end{table*}

\section{Frame-Selection Latency}
\label{app:frame-selection-latency}

Table~\ref{tab:appendix-frame-selection-latency} reports average per-instance latency for all 1,337 LongVideoBench validation instances, grouped by video duration into 498, 273, 290, 168, and 108 instances. All methods use BLIP-large on one NVIDIA GeForce RTX 3090 with a batch size of 8. The additional $\ell_2$ normalization of visual features required by \our is included in the reported latency. One-time model loading and warm-up are excluded.

The table separates video decoding, BLIP preprocessing, BLIP inference, and residual frame-selection operations, including video metadata retrieval, BLIP output post-processing, and method-specific selection overhead. Beyond five minutes, the total latency of \our remains between 10.452 and 11.354 seconds, whereas AKS and FOCUS reach 143.361 and 80.256 seconds for videos of 30--60 minutes. Because residual selection time is small for all methods, the advantage mainly comes from decoding and scoring fewer frames with BLIP as video duration increases.

\section{Additional Experimental Details}
\label{app:additional-experiments}

\paragraph{Complete configuration.}
The default configuration is $K_{\max}=32$, $K_{\min}=8$, $R_{\max}=128$, $G=512$, $n_1=64$, $\ell\in[1,38]$ seconds, $\kappa=0.005$, $\epsilon=0.05$, $\tau=2$, and $\delta=10^{-6}$. Here, $K_{\min}$ specifies the minimum number of keyframes before the final coverage-saturation test is enabled; it does not change the greedy order. The same configuration and random seed 42 are used for both benchmarks and all downstream models. Each ablation changes one component at a time.

\paragraph{Frame scoring and batching.}
The BLIP-large frame scorer reuses one image encoding to obtain the frame--query relevance score and the query-independent visual feature. The feature is $\ell_2$-normalized before coverage computation. The coarse probe and the actively acquired frames are processed in two separate batched scoring phases. Active-acquisition planning invokes no frame scorer. Accordingly, $R$ counts distinct scored frames rather than the number of minibatch forward passes.

\subsection{Parameter Sensitivity}
\label{app:sensitivity}

Table~\ref{tab:parameter-sensitivity-values} provides the exact values shown in Figure~\ref{fig:parameter-sensitivity}. Each result uses Qwen2-VL-7B-Instruct and is averaged over LVB and VMME. Only the listed parameter is changed from the default configuration.

Across these one-at-a-time sweeps, mean accuracy varies by at most 0.95 points, $\bar R$ ranges from 77.1 to 83.2, and $\overline{|\mathcal S|}$ ranges from 24.2 to 27.5. Increasing $\epsilon$ enables earlier saturation stopping and therefore reduces both frame counts. The effects of $\kappa$ and $\ell_{\max}$ are smaller within the tested ranges. 

\begin{table}[ht]
\centering
\small
\caption{Parameter sensitivity results. Asterisks mark the default values.}
\label{tab:parameter-sensitivity-values}
\begin{tabular}{@{}lcrrr@{}}
\toprule
Parameter & Value & Mean & $\bar R$ & $\overline{|\mathcal S|}$ \\
\midrule
\multirow{3}{*}{$\epsilon$}
& $0.03$ & 58.76 & 83.2 & 27.5 \\
& $0.05^{\ast}$ & \textbf{59.17} & 79.6 & 25.8 \\
& $0.07$ & 58.22 & 77.1 & 24.2 \\
\midrule
\multirow{3}{*}{$\kappa$}
& $0.001$ & 58.76 & 77.6 & 24.6 \\
& $0.005^{\ast}$ & \textbf{59.17} & 79.6 & 25.8 \\
& $0.01$ & 59.00 & 80.0 & 26.2 \\
\midrule
\multirow{3}{*}{$\ell_{\max}$}
& $34$ & 58.70 & 81.0 & 25.9 \\
& $38^{\ast}$ & \textbf{59.17} & 79.6 & 25.8 \\
& $42$ & 58.74 & 78.5 & 25.6 \\
\bottomrule
\end{tabular}
\end{table}

\subsection{Detailed Component Ablations}
\label{app:ablation-details}

Table~\ref{tab:ablation} reports the exact Qwen2-VL-7B-Instruct results underlying Figure~\ref{fig:ablation-tradeoff}. Mean, $\bar R$, and $\overline{|\mathcal S|}$ are unweighted averages over LVB and VMME, and $\Delta$ is the change in mean accuracy relative to full \our.

\begin{table*}[htbp]
\centering
\small
\caption{Detailed component ablations.}
\label{tab:ablation}
\begin{tabular}{@{}lccrrrr@{}}
\toprule
Variant & LVB & VMME & Mean & $\Delta$ & $\bar R$ & $\overline{|\mathcal S|}$ \\
\midrule
Full CSES & 58.34 & 60.00 & \textbf{59.17} & -- & 79.6 & 25.8 \\
Remove exploration $\sigma$ & 58.19 & 59.67 & 58.93 & $-0.24$ & 78.3 & 25.8 \\
Uniform 128 instead of acquisition & 58.04 & 59.04 & 58.54 & $-0.63$ & 114.1 & 26.9 \\
Remove predicted relevance $\hat r$ & 57.07 & 59.33 & 58.20 & $-0.97$ & 83.4 & 25.9 \\
Remove temporal coverage $A$ & 56.99 & 58.78 & 57.88 & $-1.29$ & 79.6 & 16.2 \\
Fix bandwidth $\ell=19$ s & 55.65 & 59.52 & 57.58 & $-1.59$ & 67.9 & 23.5 \\
Remove visual coverage $\Phi$ & 58.12 & 56.52 & 57.32 & $-1.85$ & 79.6 & 22.4 \\
Uniform 64 only (no acquisition) & 54.45 & 59.33 & 56.89 & $-2.28$ & 57.1 & 24.7 \\
\bottomrule
\end{tabular}
\end{table*}

\section{Active-Acquisition Details}
\label{app:active-acquisition-details}

Let $\mathcal O$ denote the uniformly scored probe set available at the start of Stage~2, $\mathcal G$ a set of up to $G$ distinct uniformly spaced unscored candidates, and $\mathcal P$ the candidates planned for additional scoring. The implementation includes the temporal endpoints in $\mathcal O$ whenever $n_1\ge2$. Consequently, every candidate $c\in\mathcal G$ has scored probes with the nearest timestamp before and after $t_c$, denoted by $j_{\mathrm L}(c)$ and $j_{\mathrm R}(c)$, respectively.

For each $c\in\mathcal G$, the relevance estimate $\hat r(c)$ and visual-transition cue $\mathrm{vis}(c)$ are computed from $\mathcal O$ using Equations~\eqref{eq:relevance-prediction} and~\eqref{eq:exploration}. These two quantities remain fixed during batch planning. The method instead updates virtual temporal coverage as candidates are added to $\mathcal P$:

\[
\mathrm{cov}_v(c)=\max_{j\in\mathcal O\cup\mathcal P}A_{cj}.
\]
Initially, $\mathcal P=\varnothing$, so coverage is determined only by scored probes. Adding a candidate to $\mathcal P$ updates coverage through its timestamp, without using its unknown relevance score or visual feature. At each planning step, the method recomputes
\[
\begin{aligned}
\sigma(c)&=(1-\mathrm{cov}_v(c))\mathrm{vis}(c),\\
W(c)&=u\bigl(\hat r(c)+\kappa\bigr)+(1-u)\sigma(c).
\end{aligned}
\]
For every eligible candidate $x\in\mathcal G\setminus\mathcal P$, its weighted marginal temporal-coverage gain is
\[
\Delta(x)=
\sum_{c\in\mathcal G}
W(c)\,[A_{xc}-\mathrm{cov}_v(c)]_+.
\]
Here, $[z]_+=\max(z,0)$. The maximum-gain candidate is added to $\mathcal P$, after which coverage, $\sigma$, $W$, and all subsequent gains are updated. Thus, $W(c)$ weights coverage demand over the timeline rather than serving as an independent ranking score.

Planning stops at weighted coverage saturation, when the scoring budget or candidate pool is exhausted, or when no candidate provides a positive marginal gain. The planned set is then scored in one batched phase and added to $\mathcal O$. 

\section{Submodularity of the Coverage Objective}
\label{app:submodularity-proof}

The following result concerns final coverage selection, where the scored set $\mathcal O$, the relevance scores, and the joint kernel $\mathcal K$ are fixed. Since $r_i\in[0,1]$ and $\kappa>0$, the weight $w_i=r_i+\kappa$ is positive. Moreover, $\mathcal K_{ij}\ge0$ by construction.

\noindent\textbf{Proposition 1.}
For any fixed $\mathcal O$ and $\mathcal K$, the objective
\[
F(\mathcal S)=
\sum_{i\in\mathcal O}
w_i\max_{j\in\mathcal S}\mathcal K_{ij},
\qquad
\max_{j\in\varnothing}\mathcal K_{ij}:=0,
\]
is a normalized, monotone, and submodular set function defined on all subsets of $\mathcal O$.

\noindent\textit{Proof.}
For each $i\in\mathcal O$, define
\[
g_i(\mathcal S)=\max_{j\in\mathcal S}\mathcal K_{ij},
\qquad g_i(\varnothing)=0.
\]
The empty-set convention gives $F(\varnothing)=0$. For any
$\mathcal A\subseteq\mathcal B\subseteq\mathcal O$, taking a maximum over
a larger set gives $g_i(\mathcal A)\le g_i(\mathcal B)$, so $g_i$ is
monotone. For any $x\notin\mathcal B$, its marginal gain is
\[
\begin{aligned}
g_i(\mathcal A\cup\{x\})-g_i(\mathcal A)
&=[\mathcal K_{ix}-g_i(\mathcal A)]_+\\
&\ge[\mathcal K_{ix}-g_i(\mathcal B)]_+\\
&=g_i(\mathcal B\cup\{x\})-g_i(\mathcal B),
\end{aligned}
\]
where $[z]_+=\max(z,0)$. Thus, $g_i$ satisfies diminishing returns and is
submodular. Finally, since $F=\sum_{i\in\mathcal O}w_i g_i$ and $w_i\ge0$,
the marginal gain of $F$ for any $\mathcal A\subseteq\mathcal B\subseteq
\mathcal O$ and $x\notin\mathcal B$ is a nonnegative weighted sum of the
marginal gains of the functions $g_i$:
\[
\begin{aligned}
F(\mathcal A\cup\{x\})-F(\mathcal A)
&=\sum_{i\in\mathcal O}w_i
\bigl[g_i(\mathcal A\cup\{x\})-g_i(\mathcal A)\bigr]\\
&\ge\sum_{i\in\mathcal O}w_i
\bigl[g_i(\mathcal B\cup\{x\})-g_i(\mathcal B)\bigr]\\
&=F(\mathcal B\cup\{x\})-F(\mathcal B)\ge0.
\end{aligned}
\]
The first inequality follows from the submodularity of each $g_i$, and the
final inequality follows from their monotonicity and the nonnegative weights.
Therefore, $F$ is normalized, monotone, and submodular.
\nobreak\hfill\mbox{$\square$}


\section{Coverage Selection Pseudocode}
\label{app:coverage-pseudocode}

Algorithm~\ref{alg:coverage-summary} instantiates the relevance-weighted coverage objective in Equation~\eqref{eq:coverage-objective}. It maintains
$\mathrm{cov}_i=\max_{s\in\mathcal S}\mathcal K_{is}$, with
$\mathrm{cov}_i=0$ for $\mathcal S=\varnothing$. Hence,
$g(j)=\sum_i w_i[\mathcal K_{ij}-\mathrm{cov}_i]_+$ is exactly the
marginal gain $F(\mathcal S\cup\{j\})-F(\mathcal S)$. 

Because $\mathcal K_{ij}\in[0,1]$ and $\mathcal K_{ii}=1$, selecting all
scored frames yields the full-set reference coverage
$F_{\mathrm{tot}}=F(\mathcal O)=\sum_i w_i$. After at least
$K_{\min}$ frames have been selected, the algorithm stops when the
accumulated coverage reaches $(1-\epsilon)F_{\mathrm{tot}}$. Otherwise,
it continues until it reaches $K_{\max}$, exhausts the scored set, or
finds no positive marginal gain. Concentrated evidence may therefore
reach saturation with a small representative set, whereas evidence
distributed across the video may require more keyframes but remains
bounded by $K_{\max}$. Sorting the returned set by timestamp also leaves
the coverage objective unchanged.

\begin{algorithm}[!t]
\caption{Greedy relevance-weighted coverage selection.}
\label{alg:coverage-summary}
\footnotesize
\KwIn{Scored set $\mathcal O$ with $\{t_i,r_i,e_i\}_{i\in\mathcal O}$; bandwidth $\ell$; $\kappa,K_{\min},K_{\max},\epsilon$}
\KwOut{Selected keyframe set $\mathcal S\subseteq\mathcal O$}
$\mathcal K_{ij}\leftarrow\max(\langle e_i,e_j\rangle,0)
\exp\!\left(-\frac{(t_i-t_j)^2}{2\ell^2}\right)$;
$\mathcal K_{ii}\leftarrow1$\;
$w_i\leftarrow r_i+\kappa$ for every $i\in\mathcal O$\;
$F_{\mathrm{tot}}\leftarrow\sum_{i\in\mathcal O}w_i$\;
$\mathcal S\leftarrow\varnothing$;
$\mathrm{cov}_i\leftarrow0$ for every $i\in\mathcal O$\;
\While{$|\mathcal{S}| < K_{\max}$}{
  \If{$\sum_{i} w_i\,\mathrm{cov}_i \ge (1-\epsilon)\,F_{\mathrm{tot}}$ \textbf{\emph{and}} $|\mathcal{S}|\ge K_{\min}$}{
    break;
  }
  $g(j)\leftarrow\sum_{i\in\mathcal{O}} w_i\,[\mathcal{K}_{ij}-\mathrm{cov}_i]_+$, $\ \forall j\in\mathcal{O}\setminus\mathcal{S}$\; 
  $j^\ast\leftarrow\arg\max_{j} g(j)$, break ties by minimizing
$\max_{s\in\mathcal S}\mathcal K_{js}$, thereby selecting the candidate farthest from $\mathcal S$\;
  \lIf{$g(j^\ast)\le 0$}{break}
  $\mathcal{S}\leftarrow\mathcal{S}\cup\{j^\ast\}$;\quad $\mathrm{cov}_i\leftarrow\max(\mathrm{cov}_i,\mathcal{K}_{ij^\ast}),\ \forall i$\;
}
\Return $\mathcal S$ sorted by timestamp\;
\end{algorithm}


\end{document}